\documentclass[9pt,twocolumn,twoside]{pnas-new}

\templatetype{pnasresearcharticle} 

\title{On a `Two Truths' Phenomenon in Spectral Graph Clustering}

\author[a,b,c,1]{Carey E.\ Priebe}
\author[b]{Youngser Park}
\author[b,d]{Joshua T. Vogelstein}
\author[e]{John M. Conroy}
\author[c,f]{Vince Lyzinski}
\author[a]{Minh Tang} 
\author[a]{Avanti Athreya} 
\author[a]{Joshua Cape} 
\author[b,g]{Eric Bridgeford}

\affil[a]{Department of Applied Mathematics and Statistics, Johns Hopkins University, Baltimore, MD 21218}
\affil[b]{Center for Imaging Science, Johns Hopkins University, Baltimore, MD 21218}
\affil[c]{Human Language Technology Center of Excellence, Johns Hopkins University, Baltimore, MD 21218}
\affil[d]{Department of Biomedical Engineering, Johns Hopkins University, Baltimore, MD 21218}
\affil[e]{Institute for Defense Analyses, Center for Computing Sciences, Bowie, MD 20715}
\affil[f]{Department of Mathematics and Statistics, University of Massachusetts, Amherst, MA 01003}
\affil[g]{Department of Biostatistics, Johns Hopkins University, Baltimore, MD 21218}

\leadauthor{Priebe} 

\significancestatement{Spectral graph clustering -- clustering the vertices of a graph based on their spectral embedding -- is of significant current interest,
finding applications throughout the sciences. But as with clustering in general, what a particular methodology identifies as ``clusters'' is defined (explicitly, or, more often, implicitly) by the clustering algorithm itself.  We provide a clear and concise demonstration of a `Two Truths' phenomenon for spectral graph clustering in which the first step -- spectral embedding -- is either Laplacian Spectral Embedding (LSE) wherein one decomposes the normalized Laplacian of the adjacency matrix, or Adjacency Spectral Embedding (ASE) given by a decomposition of the adjacency matrix itself. The two resulting clustering methods identify fundamentally different (true and meaningful) structure. 
}

\authorcontributions{Author contributions: CEP,JTV,VL,JMC conceived these ideas;
YP did the computer work; 
JTV,EB provided the connectomes; CEP,VL,MT,AA,JC developed the theory; CEP wrote this manuscript.
}

\authordeclaration{The authors declare no conflict of interest.}
\correspondingauthor{\textsuperscript{1}To whom correspondence should be addressed. E-mail: cep@jhu.edu}


\begin{abstract}
Clustering is concerned with coherently grouping observations without any explicit concept of true groupings. Spectral graph clustering -- clustering the vertices of a graph based on their spectral embedding -- is commonly approached via K-means (or, more generally, Gaussian mixture model) clustering composed with either Laplacian or Adjacency spectral embedding (LSE or ASE).
Recent theoretical results provide new understanding of the problem and solutions, and lead us to a `Two Truths' LSE vs.\ ASE spectral graph clustering phenomenon
convincingly illustrated here via a diffusion MRI connectome data set:
 the different embedding methods yield different clustering results,
 with LSE capturing left hemisphere/right hemisphere affinity structure
 and ASE capturing gray matter/white matter core-periphery structure.
\end{abstract}

\dates{This manuscript was compiled on \today}
\doi{\url{www.pnas.org/cgi/doi/10.1073/pnas.XXXXXXXXXX}}

\begin{document}

\verticaladjustment{-2pt}

\maketitle
\thispagestyle{firststyle}
\ifthenelse{\boolean{shortarticle}}{\ifthenelse{\boolean{singlecolumn}}{\abscontentformatted}{\abscontent}}{}


\dropcap{T}he purpose of this paper is to
cogently present a `Two Truths' phenomenon in spectral graph clustering,
to understand this phenomenon from a theoretical and methodological perspective,
and to demonstrate the phenomenon in a real-data case consisting of multiple graphs each with multiple categorical vertex class labels.
 
A graph or network consists of a collection of vertices or nodes $V$ representing $n$ entities together with edges or links $E$ representing the observed subset of the ${\binom{n}{2}}$ possible pairwise relationships between these entities.  Graph clustering, often associated with the concept of `community detection', is concerned with partitioning the vertices into coherent groups or clusters. By its very nature, such a partitioning must be based on 
connectivity patterns.

\begin{figure}[htbp]
\centering
\includegraphics[width=0.92\linewidth]{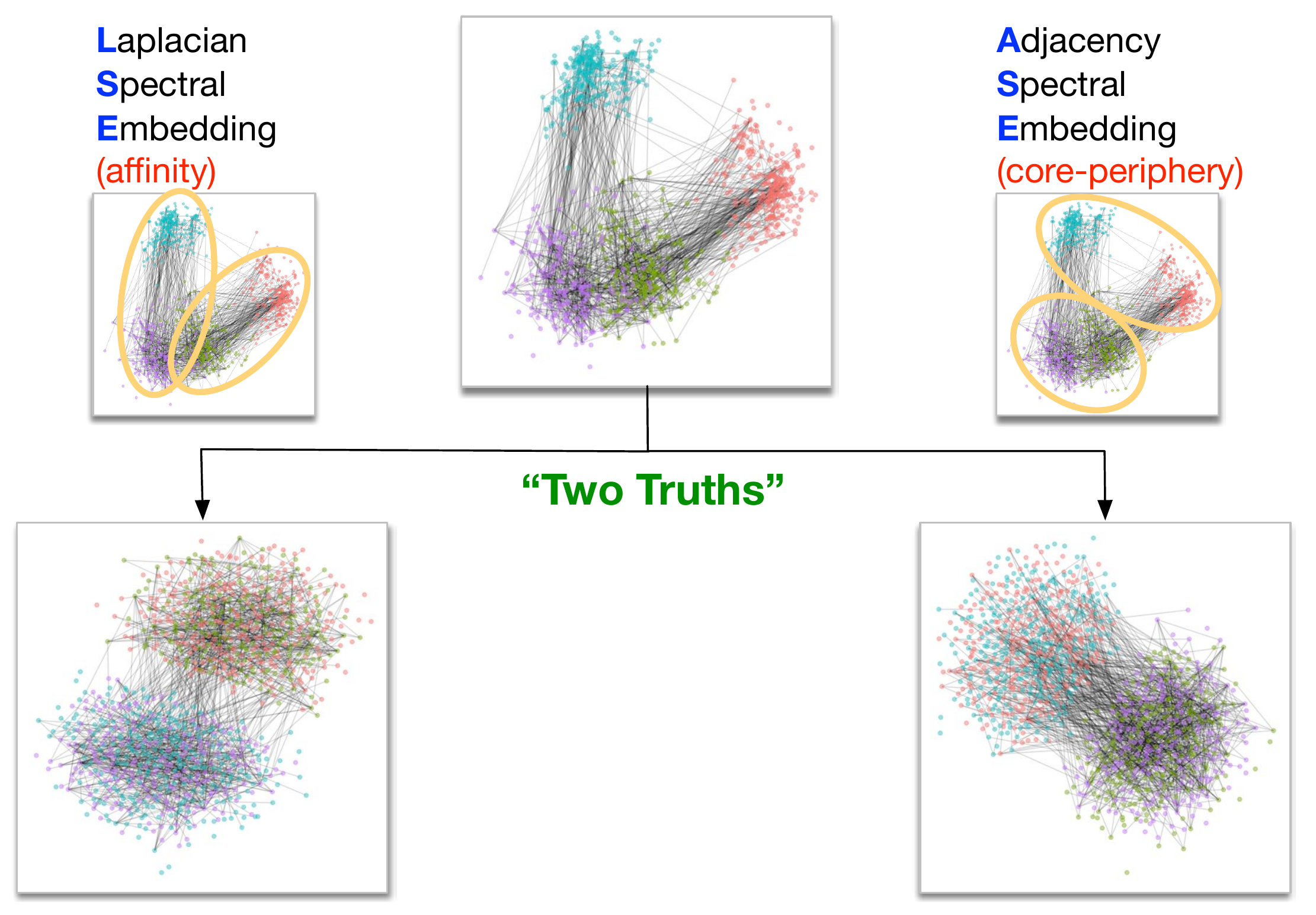}
\caption{
A `Two Truths' graph (connectome) depicting connectivity structure such that one grouping of the vertices yields affinity structure (e.g.\ left hemisphere/right hemisphere) and the other grouping yields core-periphery structure (e.g.\ gray matter/white matter).
Top center: the graph with four vertex colors.
Top left / top right: LSE groups one way; ASE groups another way.
Bottom left: the LSE truth is two densely connected groups, with sparse interconnectivity between them (affinity structure).
Bottom right: the ASE truth is one densely connected group, with sparse interconnectivity between it and the other group and sparse interconnectivity within the other group (core-periphery structure).
This paper demonstrates the `Two Truths' phenomenon illustrated in this figure -- that LSE and ASE find fundamentally different but equally meaningful network structure -- via theory, simulation, and real data analysis.
}
\label{fig:killer}
\end{figure}

It is often the case that practitioners cluster the vertices of a graph -- say, via $K$-means clustering composed with Laplacian spectral embedding -- and pronounce the method as either having performed well or poorly based on whether the resulting clusters correspond well or poorly with some known or preconceived notion of ``correct'' clustering. Indeed, such a procedure may be employed to compare two clustering methods, and to pronounce that one works better (on the particular data under consideration). However, clustering is inherently ill-defined, as there may be multiple meaningful groupings, and two clustering methods that perform differently with respect to one notion of truth may in fact be identifying inherently different, but perhaps complementary, underlying structure.
With respect to graph clustering,
\cite{Peele1602548} 
shows
that there can be no algorithm that is optimal for 
all possible community detection tasks. 
See Figure \ref{fig:killer}.

We compare and contrast Laplacian and Adjacency spectral embedding as the first step in spectral graph clustering, and demonstrate that the two methods, and the two resulting clusterings, identify different -- but both meaningful -- graph structure. We trust that this simple, clear explication will contribute to an awareness that connectivity-based structure discovery via spectral graph clustering should consider both Laplacian and Adjacency spectral embedding, and to the development of new methodologies based on this awareness.

\section*{Spectral Graph Clustering}

Given a simple graph $G=(V,E)$ on $n$ vertices, consider the associated $n \times n$ adjacency matrix $A$ in which $A_{ij}$ = 0 or 1 encoding whether vertices $i$ and $j$ in $V$ share an edge $(i,j)$ in $E$. For our simple undirected, unweighted, loopless
case, $A$ is binary with $A_{ij} \in \{0,1\}$, symmetric with $A=A^\top$, and hollow with $diag(A)=\vec{0}$.

The first step of spectral graph clustering \cite{vonLuxburg2007,rohe2011} involves embedding the graph into Euclidean space via 
an eigendecomposition.
We consider two options:
Laplacian Spectral Embedding (LSE) wherein we decompose the normalized Laplacian of the adjacency matrix, and Adjacency Spectral Embedding (ASE) given by a decomposition of the adjacency matrix itself.
With target dimension $d$, either spectral embedding method produces $n$ points in $\Re^d$, denoted by the $n \times d$ matrix $X$. 
ASE employs the 
eigendecomposition
to represent the adjacency matrix via $A = USU^{\top}$ and chooses the top $d$ 
eigenvalues by magnitude
and their associated vectors to embed the graph 
via the scaled 
eigenvectors $U_d |S_d|^{1/2}$.
Similarly, LSE embeds the graph 
via the top scaled 
eigenvectors 
of the normalized Laplacian
$\mathcal{L}(A) = D^{-1/2} A D^{-1/2}$,
where $D$ is the diagonal matrix of vertex degrees.
In either case, each vertex is mapped to the corresponding row of 
$X = U_d |S_d|^{1/2}$.

Spectral graph clustering concludes via classical Euclidean clustering of $X$. As described below, Central Limit Theorems for spectral embedding of the 
(sufficiently dense)
Stochastic Block Model via either LSE or ASE suggest Gaussian Mixture Modeling (GMM) for this clustering step.
Thus we consider spectral graph clustering to be GMM composed with LSE or ASE:
$$ \mbox{GMM} ~ \circ ~ \{\mbox{LSE},\mbox{ASE}\}. $$

\section*{Stochastic Block Model}

The random graph model we use to illustrate our phenomenon is the
Stochastic Block Model (SBM), introduced in \cite{holland1983stochastic}.
This model is parameterized by ({\it i}) a block membership
probability vector $\vec{\pi} = [\pi_1,\dots,\pi_K]^\top$ in the unit simplex
and ({\it ii}) a symmetric $K \times K$
block connectivity probability matrix $B$
with entries in $[0,1]$ governing the probability
of an edge between vertices given their block memberships.
Use of the SBM is ubiquitous in theoretical, methodological, and practical graph investigations, %
and SBMs have been shown to be universal approximators for exchangeable random graphs \cite{Olhede14722}.

For sufficiently dense graphs, both
LSE and ASE have a Central Limit Theorem \cite{athreya2013limit,tang_lse,PRD-GRDPG} demonstrating that, for large $n$, embedding via the top $d$ eigenvectors from a 
rank $d$ $K$-block SBM ($d \equiv rank(B) \leq K$) yields $n$ points in $\Re^d$ behaving approximately as a random sample from a mixture of $K$ Gaussians.
That is, given that the $i${th} vertex belongs to block $k$, the $i${th} row of $X = U_d S_d^{1/2}$ will be approximately distributed as a multivariate normal with parameters specific to block $k$, $X_i \sim \mathcal{MVN}(\mu_k,\Sigma_k)$.
The structure of the covariance matrices suggest that the GMM is called for, as an appropriate generalization of $K$-means clustering.
Therefore, GMM$(X)$ via Maximum Likelihood will produce mixture parameter estimates and associated asymptotically perfect clustering, using either LSE or ASE.
For finite $n$, however, LSE and ASE yield different clustering performance, and neither dominates the other.

We will make significant conceptual use of the positive definite 2-block SBM ($K=2$), with
$$
B =
\begin{bmatrix}
 B_{11} & B_{12} \\
 B_{21} & B_{22} \\
\end{bmatrix}
=
\begin{bmatrix}
 a & b \\
 b & c \\
\end{bmatrix}
$$
which henceforth we shall abbreviate as $B = [a,b;b,c]$.
In this simple setting, two general/generic cases present themselves: affinity and core-periphery.

Affinity: $a,c \gg b$. 
An SBM with $B = [a,b;b,c]$ is said to exhibit affinity structure if
each of the two blocks have a relatively high within-block connectivity probability compared to the between-block connectivity probability.

Core-periphery: $a \gg b,c$. 
An SBM with $B = [a,b;b,c]$ is said to exhibit core-periphery structure if
one of the two blocks has a relatively high within-block connectivity probability compared to both the other block's within-block connectivity probability and the between-block connectivity probability.

The relative performance of LSE and ASE for these two cases provides the foundation for our analyses.
Informally: LSE outperforms ASE for affinity, and ASE is the better choice for core-periphery.
We make this clustering performance assessment analytically precise via Chernoff Information, and we demonstrate this in practice via Adjusted Rand Index.

\section*{Clustering Performance Assessment}

We consider two approaches to assessing the performance of a given clustering, defined to be a partition of $[n] \equiv \{1,\dots,n\}$ into a disjoint union of $K$ partition cells or clusters.
For our purposes --
demonstrating a `Two Truths' phenomenon in LSE vs.\ ASE spectral graph clustering --
we will consider the case in which 
there is a `true' or meaningful clustering of the vertices
against which we can assess performance,
but we emphasize that in practice such a truth is neither known nor necessarily unique.

\subsection*{Chernoff Information}
Comparing and contrasting the relative performance of LSE vs.\ ASE via the concept of Chernoff information
\cite{chernoff_1952,chernoff_1956},
in the context of their respective CLTs, provides a limit theorem notion of superiority.
Thus, in the SBM, we allude to the GMM provided by the CLT for either LSE or ASE.

The Chernoff information between two distributions is the exponential rate at which the decision-theoretic Bayes error
decreases as a function of sample size.
In the 2-block SBM, with the true clustering of the vertices given by the block memberships, we are interested in the large-sample optimal error rate for recovering the underlying block memberships after the spectral embedding step has been carried out.
Thus we require
the Chernoff information $C(F_1,F_2)$ when $F_1 =  \mathcal{MVN}(\mu_1, \Sigma_1)$ and $F_2 = \mathcal{MVN}(\mu_2, \Sigma_2)$ are multivariate normals.
Letting $\Sigma_t = t \Sigma_1 + (1 - t) \Sigma_2$
and
\begin{align*}
h(t;F_1,F_2) = \ & \frac{t(1 - t)}{2} (\mu_1 - \mu_2)^{\top}\Sigma_t^{-1}(\mu_1 - \mu_2) \\ & + \frac{1}{2} \log \frac{|\Sigma_t|}{|\Sigma_1|^{t} |\Sigma_2|^{1 - t}}
\end{align*}
we have
$$
\rho_{F_1, F_2} = \sup_{t \in (0,1)} h(t;F_1,F_2).
$$
This provides both $\rho_L$ and $\rho_A$ when using the large-sample GMM parameters for $F_1, F_2$ obtained from the LSE and ASE embeddings, respectively, for a particular 2-block SBM distribution (defined by its
block membership
probability vector $\vec{\pi}$ and block connectivity probability matrix $B$).
We will make use of the Chernoff ratio $\rho = \rho_A/\rho_L$; $\rho > 1$ implies ASE is preferred while $\rho < 1$ implies LSE is preferred.
(Recall that as the Chernoff information increases, the large-sample optimal error rate decreases.)
Chernoff analysis in the 2-block SBM demonstrates that, in general, LSE is preferred for affinity while ASE is preferred for core-periphery \cite{tang_lse,JCapeCC}.


\subsection*{Adjusted Rand Index}
In practice, we wish to empirically assess the performance of a particular clustering algorithm
on a given graph.
There are numerous cluster assessment criteria available in the literature:
  Rand Index (RI) \cite{hubert85},
  Normalized Mutual Information (NMI) \citep{dadiduar05},
  Variation of Information (VI) \citep{me07},
  Jaccard \citep{ja1912},
  etc.
These are typically employed to compare either an empirical clustering against a `truth', or two separate empirical clusterings.
For concreteness, we consider the well known Adjusted Rand Index (ARI), popular in machine learning, which normalizes RI so that
expected chance performance is zero: ARI is the adjusted-for-chance probability that two partitions of a collection of data points will agree for a randomly chosen pair of data points, putting the pair into the same partition cell in both clusterings, or splitting the pair into different cells in both clusterings.
(Our empirical connectome results are essentially unchanged when using other cluster assessment criteria.)

In the context of spectral clustering via $\mbox{GMM} ~ \circ ~ \{\mbox{LSE},\mbox{ASE}\}$,
we consider $\mathcal{C}_{LSE}$ and $\mathcal{C}_{ASE}$ to be the two clusterings of the vertices of a given graph.
Then ARI($\mathcal{C}_{LSE}$,$\mathcal{C}_{ASE}$)
assesses their agreement: 
ARI($\mathcal{C}_{LSE}$,$\mathcal{C}_{ASE}$) $= 1$ implies that the two clusterings are identical;
ARI($\mathcal{C}_{LSE}$,$\mathcal{C}_{ASE}$) $\approx 0$ implies that the two spectral embedding methods are ``operationally orthogonal.'' (Significance is assessed via permutation testing.)

In the context of `Two Truths',
we consider $\mathcal{C}_1$ and $\mathcal{C}_2$ to be two known `true' or meaningful clusterings of the vertices.
Then, with 
$\mathcal{C}_{SE}$ being either $\mathcal{C}_{LSE}$ or $\mathcal{C}_{ASE}$,
ARI($C_{SE}$,$\mathcal{C}_1$)
$\gg$
ARI($C_{SE}$,$\mathcal{C}_2$)
implies that the spectral embedding method under consideration is more adept at discovering truth $\mathcal{C}_1$ than truth $\mathcal{C}_2$.
Analogous to the theoretical Chernoff analysis, ARI simulation studies in the 2-block SBM demonstrate that, in general, LSE is preferred for affinity while ASE is preferred for core-periphery.


\section*{Model Selection $\times$ 2}

In order to perform the spectral graph clustering $ \mbox{GMM} ~ \circ ~ \{\mbox{LSE},\mbox{ASE}\}$ in practice, we must address two inherent model selection problems:
we must choose the embedding dimension ($\widehat{d}$) and the number of clusters ($\widehat{K}$).

\subsection*{SBM vs.\ network histogram}

If the SBM model were actually true, then as $n \to \infty$ any reasonable procedure for estimating the SVD rank would yield a consistent estimator $\widehat{d} \to d$ and any reasonable procedure for estimating the number of clusters would yield a consistent estimator $\widehat{K} \to K$.
Critically, the universal approximation result of \cite{Olhede14722} shows that SBMs provide a principled `network histogram' model even without the assumption that the SBM model with some fixed $(d,K)$ actually holds. Thus, practical model selection for spectral graph clustering is concerned with choosing ($\widehat{d},\widehat{K}$) so as to provide a useful approximation.

The bias-variance tradeoff demonstrates that
any quest for a universally optimal methodology for choosing the ``best'' dimension and number of clusters, in general, for finite $n$,
is a losing proposition.
Even for a low-rank model, subsequent inference may be optimized by choosing a dimension {\em smaller than} the true signal dimension, and even for a mixture of $K$ Gaussians, inference performance may be optimized by choosing a number of clusters {\em smaller than} the true cluster complexity.
In the case of semiparametric SBM fitting, wherein low-rank and finite mixture are employed as a practical modeling convenience as opposed to a believed true model, and one presumes that both $\widehat{d}$ and $\widehat{K}$ will tend to infinity as $n \to \infty$, these bias-variance tradeoff considerations are exacerbated.

For $\widehat{d}$ and $\widehat{K}$ below, we make principled methodological choices for simplicity and concreteness, but make no claim that these are best in general or even for the connectome data considered herein.
Nevertheless, one must choose an embedding dimension and a mixture complexity, and thus we proceed.

\subsection*{Choosing the embedding dimension $\widehat{d}$}

A ubiquitous and principled general methodology for choosing the number of dimensions in eigendecompositions and SVDs
(e.g., principal components analysis, factor analysis, spectral embedding, etc.)\
  is to examine the so-called scree plot
  and look for ``elbows'' 
defining the cut-off between the top (signal) dimensions and the noise dimensions.
There are a plethora of variations for automating this singular value thresholding (SVT);
Section 2.8 of \cite{Jackson} provides a comprehensive discussion in the context of principal components,
and 
\cite{chatterjee2015}
 provides a theoretically-justified (but perhaps practically suspect, for small $n$) universal SVT.
We consider the profile likelihood SVT method of \cite{Zhu:2006fv}.
Given $A = USU^{\top}$ (for either LSE or ASE)
 the singular values $S$ are used to choose the embedding dimension $\widehat{d}$ via
$$\widehat{d} = \arg\max_{d} ProfileLikelihood_{S}(d)$$
where $ProfileLikelihood_{S}(d)$ provides a definition for the magnitude of the ``gap'' after the first $d$ singular values.


\subsection*{Choosing the number of clusters $\widehat{K}$}

Choosing the number of clusters in Gaussian mixture models is most often addressed by maximizing a fitness criterion penalized by model complexity.
Common approaches include
 Akaike Information Criterion (AIC) \citep{akaike1974new},
 Bayesian Information Criterion (BIC) \citep{BIC},
 Minimum Description Length (MDL) \citep{MDL},
 etc.
We consider penalized likelihood via BIC \citep{mclust2012}.
Given $n$ points in $\Re^d$ represented by $X = U_d S_d^{1/2}$ (obtained via either LSE or ASE) and letting $\theta_K$ represent the GMM parameter vector whose dimension $dim(\theta_K)$ is a function of the data dimension $d$, the mixture complexity $\widehat{K}$ is chosen via
$$\widehat{K} = \arg\max_K PenalizedLikelihood_X(\widehat{\theta}_K)$$
where $PenalizedLikelihood_X(\widehat{\theta}_K)$ is twice the log-likelihood of the data $X$ evaluated at the GMM with mixture parameter estimate $\widehat{\theta}_K$ penalized by $dim(\theta_K) \cdot \ln n$.
For spectral clustering, we employ BIC for $\widehat{K}$ after spectral embedding, so $X \in \Re^{\widehat{d}}$ with $\widehat{d}$ chosen as above.


\section*{Connectome Data}

\begin{figure}
\centering
\includegraphics[width=1.0\linewidth]{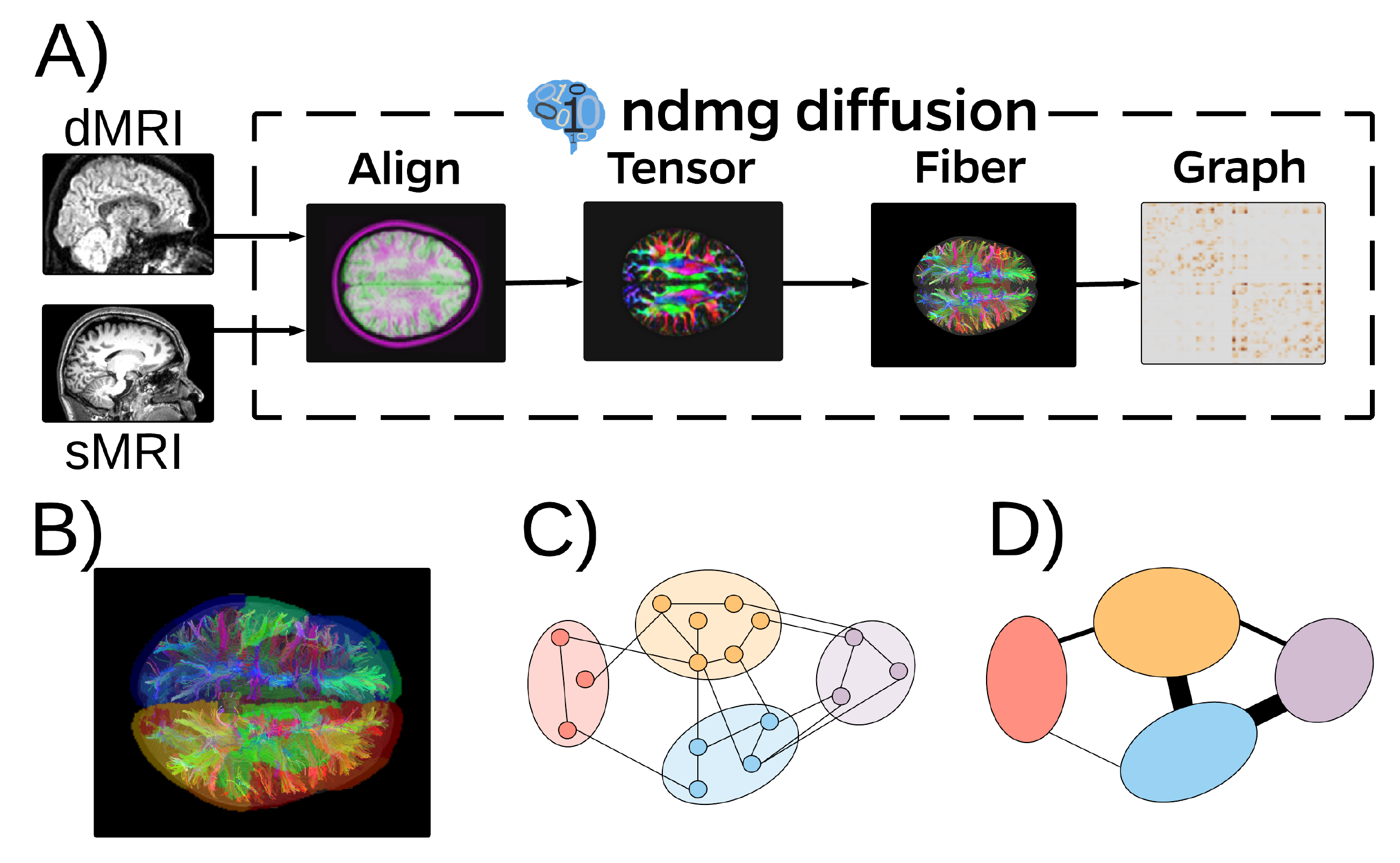}
\caption{Connectome data generation.
The output is diffusion MRI graphs on $\approx$1M vertices.
Spatial vertex contraction yields graphs on $\approx$70K vertices from which we extract largest connected components of $\approx$40K vertices
with
$\{\mbox{Left,Right}\}$ and $\{\mbox{Gray,White}\}$
labels for each vertex.
Figure \ref{fig:killer} depicts (a subsample from) one such graph.
}
\label{fig:DataGen}
\end{figure}

We consider for illustration
a diffusion MRI data set consisting of
$114$
connectomes
(57 subjects, 2 scans each)
with 
72,783
vertices each and both Left/Right/other hemispheric and Gray/White/other tissue attributes for each vertex.
Graphs were estimated using the NDMG pipeline 
\cite{Kiar188706},
with vertices representing sub-regions defined via spatial proximity and edges by tensor-based fiber streamlines connecting
these regions.
See Figure \ref{fig:DataGen}.


The actual graphs we consider are the largest connected component (LCC) of the induced subgraph on the vertices labeled as both Left or Right and Gray or White.
This yields $m=114$ connected graphs on $n \approx 40,000$ vertices.
Additionally, for each graph every vertex has a $\{\mbox{Left,Right}\}$ label and a $\{\mbox{Gray,White}\}$ label, which we sometimes find convenient to consider as a single label in $\{\mbox{LG,LW,RG,RW}\}$.

\subsection*{Sparsity}
The only notions of sparsity relevant here are linear algebraic: whether there are enough edges in the graph to support spectral embedding, and are there few enough to allow for sparse matrix computations.  We have a collection of observed connectomes and we want to cluster the vertices in these graphs, as opposed to in an unobserved sequence with the number of vertices tending to infinity.  Our connectomes have, on average, $n \approx 40,000$ vertices and $e \approx 2,000,000$ edges, for an average degree $2e/n \approx 100$ and a graph density $e/\binom{n}{2} \approx 0.0025$.


\section*{Synthetic Analysis}

\begin{figure}[tbhp]
\centering
\includegraphics[width=1.0\linewidth]{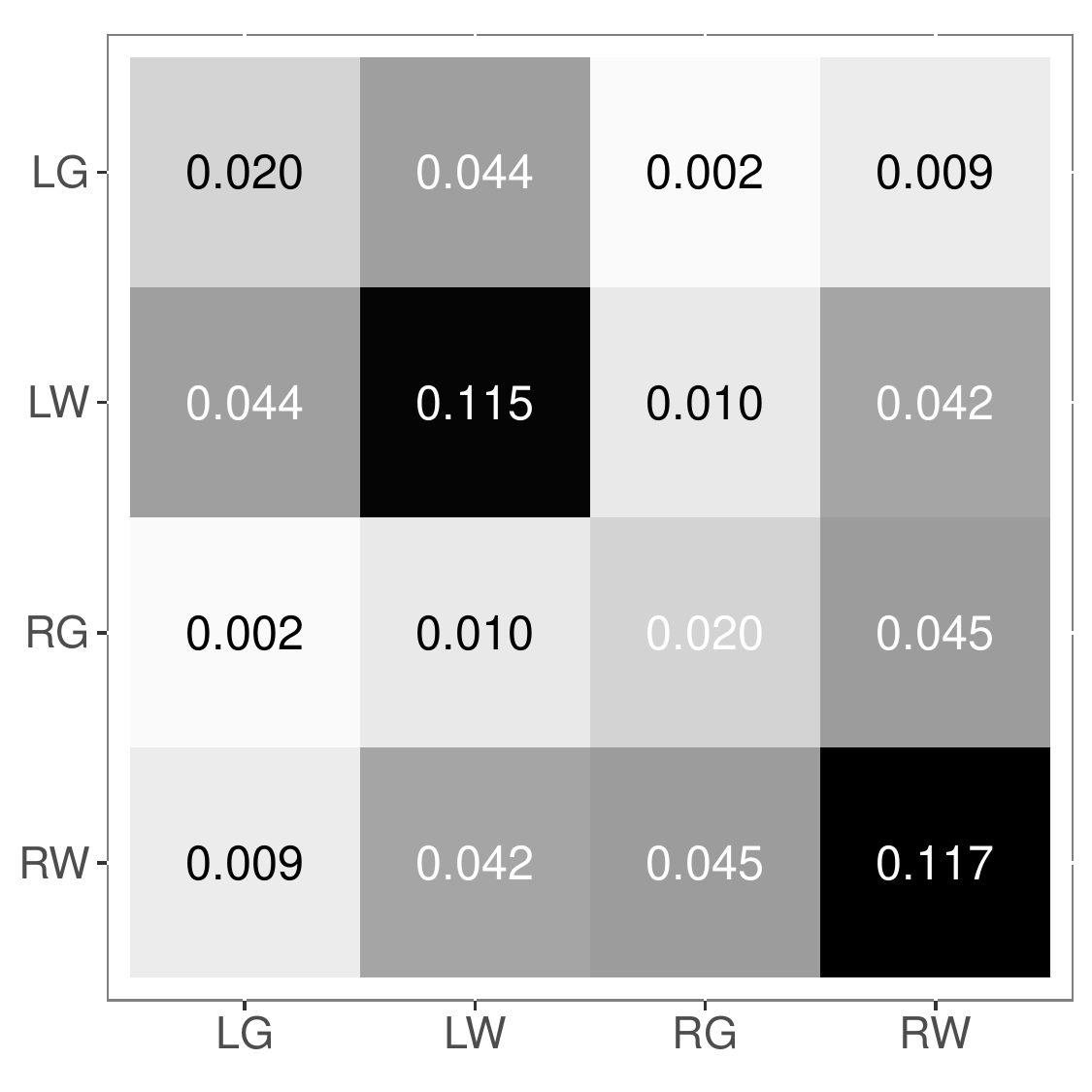}
\caption{Block connectivity probability matrix for the $\{\mbox{LG,LW,RG,RW}\}$ a priori projection of the composite connectome onto the 4-block SBM.
The two two-block projections ($\{\mbox{Left},\mbox{Right}\}$ \& $\{\mbox{Gray},\mbox{White}\}$) are shown in Figure \ref{fig:LRandGW}.
This synthetic SBM exhibits the `Two Truths' phenomenon both theoretically (via Chernoff analysis) and in simulation (via Monte Carlo).
}
\label{fig:LRGW}
\end{figure}

We consider a synthetic data analysis via a priori projections onto the SBM -- block model estimates based on known or assumed block memberships.
Averaging the collection of $m=114$ connectomes yields the composite (weighted) graph adjacency matrix $\bar{A}$. 
The $\{\mbox{LG,LW,RG,RW}\}$ projection of
the binarized
$\bar{A}$ onto the 4-block SBM yields the block connectivity probability matrix $B$ presented in Figure \ref{fig:LRGW} and the block membership probability vector $\vec{\pi} = [0.28, 0.22, 0.28, 0.22]^\top$.
Limit theory demonstrates that spectral graph clustering 
using $d=K=4$ will, for large $n$, correctly identify block memberships for this 4-block case when using either LSE or ASE.
Our interest is to compare and contrast the two spectral embedding methods for clustering into 2 clusters.
We demonstrate that this synthetic 
case exhibits the `Two Truths' phenomenon both theoretically and in simulation 
--
the $\{\mbox{LG,LW,RG,RW}\}$ a priori projection of our composite connectome
yields
a 4-block 
`Two Truths' SBM.

\begin{figure}[tbhp]
\centering
\includegraphics[width=.45\linewidth]{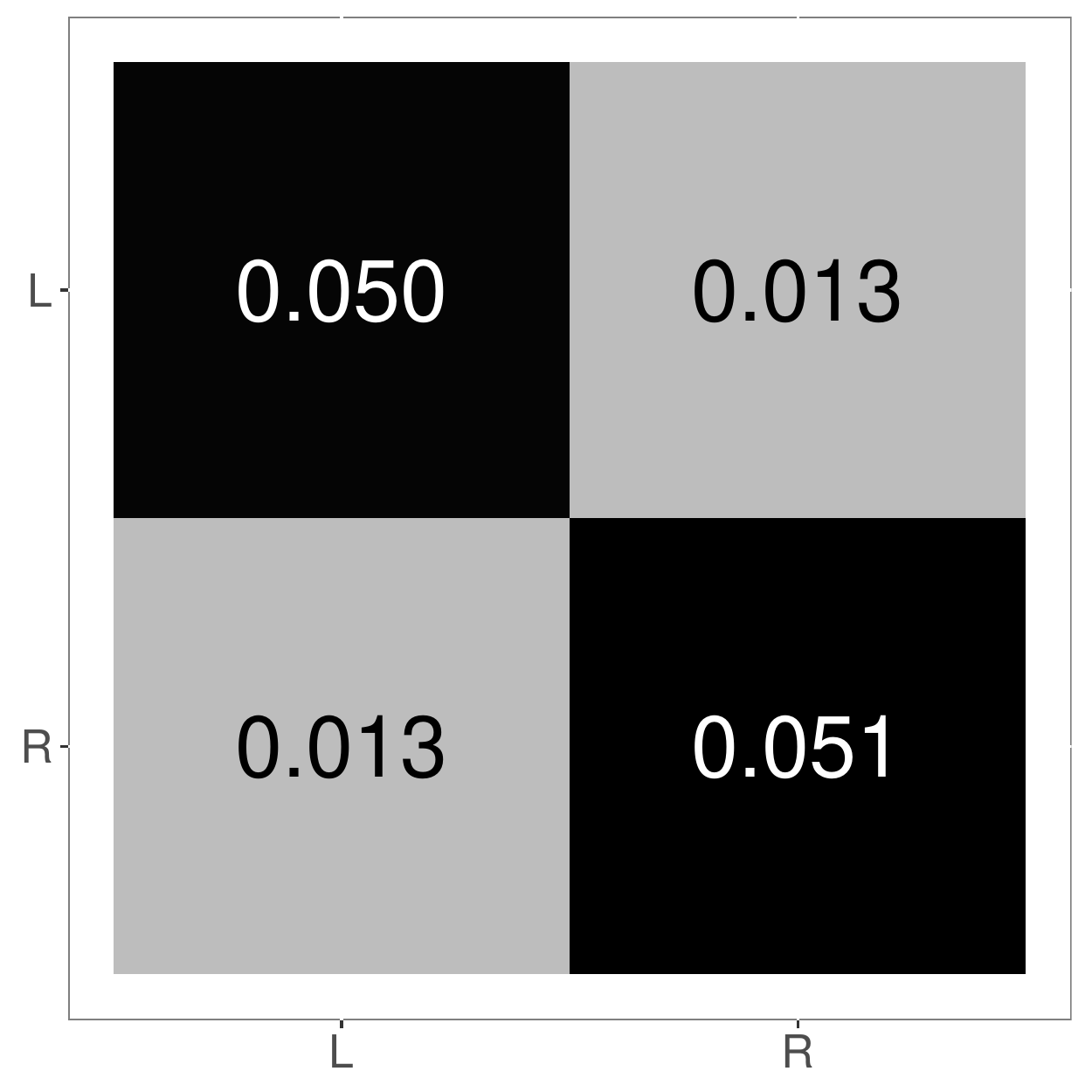}
\includegraphics[width=.45\linewidth]{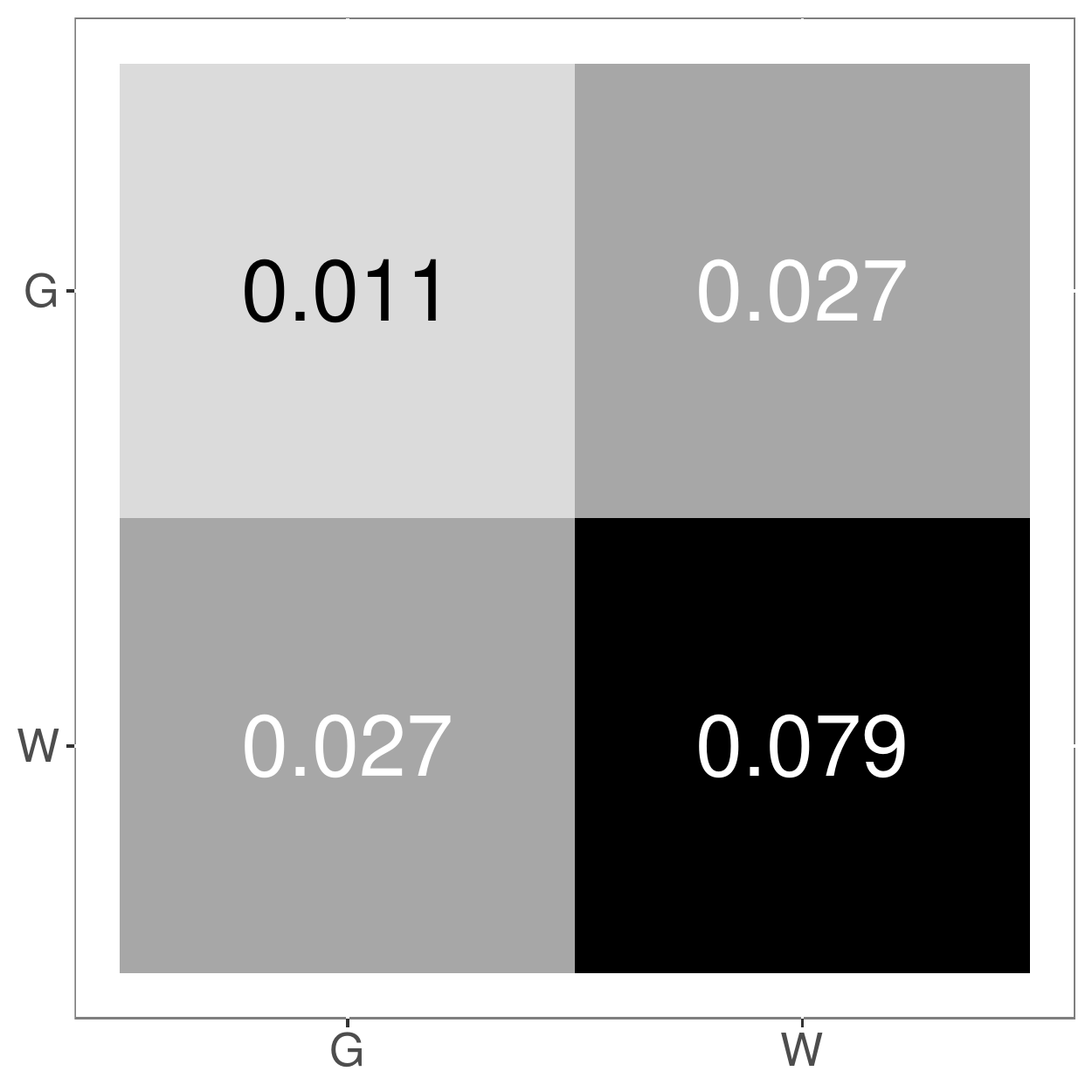}
\caption{Block connectivity probability matrices for the a priori projection of the composite connectome onto the 2-block SBM for (left panel) $\{\mbox{Left},\mbox{Right}\}$ \& (right panel) $\{\mbox{Gray},\mbox{White}\}$.
$\{\mbox{Left},\mbox{Right}\}$ exhibits affinity structure, with Chernoff ratio < 1; $\{\mbox{Gray},\mbox{White}\}$ exhibits core-periphery structure, with Chernoff ratio > 1.
}
\label{fig:LRandGW}
\end{figure}

\subsection*{2-Block Projections}

A priori projections 
onto the 2-block SBM for $\{\mbox{Left,Right}\}$ and $\{\mbox{Gray,White}\}$ yield the two block connectivity probability matrices 
shown in 
Figure \ref{fig:LRandGW}.
It is apparent that the $\{\mbox{Left,Right}\}$ a priori block connectivity probability matrix $B=[a,b;b,c]$ represents an affinity SBM with $a \approx c \gg b$
and the $\{\mbox{Gray,White}\}$ a priori projection yields a core-periphery SBM with $c \gg a \approx b$.
It remains to investigate the extent to which the
Chernoff analysis from the 2-block setting (LSE is preferred for affinity while ASE is preferred for core-periphery)
extends to such a 4-block `Two Truths' case;
we do so theoretically and in simulation using this synthetic model 
derived from the $\{\mbox{LG,LW,RG,RW}\}$ a priori projection of our composite connectome
in the next two subsections, and then empirically on the original connectomes in the following section.

\subsection*{Theoretical Results}

Analysis using the large-sample Gaussian mixture model approximations from the LSE and ASE CLTs
shows that the
2-dimensional embedding of the 4-block model,
when clustered into 
2 
clusters,
will yield
 \{ \{LG,LW\} , \{RG,RW\} \} (i.e., \{Left,Right\})
 when embedding via LSE
and
 \{ \{LG,RG\} , \{LW,RW\} \} (i.e., \{Gray,White\})
 when using ASE.
That is, 
using numerical integration 
for the $d=K=2$ $\mbox{GMM} ~ \circ ~ \mbox{LSE}$, the largest 
Kullback-Leibler divergence (as a surrogate for Chernoff information)
among the 10 possible ways of grouping the 4 Gaussians into two clusters is for the \{~\{LG,LW\}~,~\{RG,RW\}~\} grouping,
and the largest of these 
values for the $\mbox{GMM} ~ \circ ~ \mbox{ASE}$ is for the \{~\{LG,RG\}~,~\{LW,RW\}~\} grouping.

\subsection*{Simulation Results}

We augment the Chernoff limit theory via Monte Carlo simulation, sampling graphs from the 4-block model
and running the
$\mbox{GMM} ~ \circ ~ \{\mbox{LSE},\mbox{ASE}\}$
algorithm specifying $\widehat{d}=\widehat{K}=2$.
This 
results in LSE finding $\{\mbox{Left},\mbox{Right}\}$ (ARI > 0.95) with probability > 0.95 and ASE finding $\{\mbox{Gray},\mbox{White}\}$ (ARI > 0.95) with probability > 0.95.


\section*{Connectome Results}

\begin{figure}
\centering
\includegraphics[width=\linewidth]{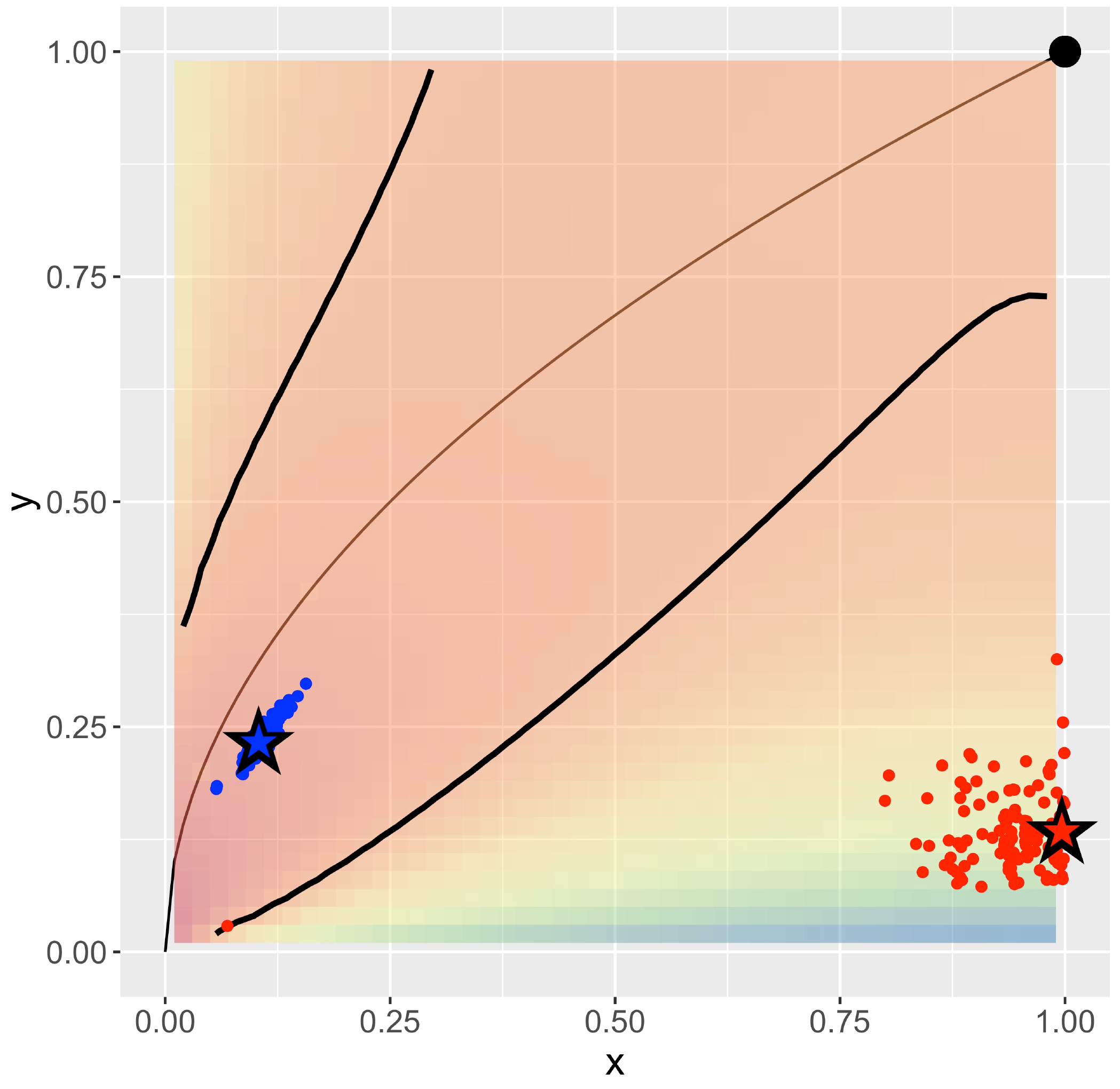}
\caption{For each of our 114 connectomes, we plot the a priori 2-block SBM projections for $\{\mbox{Left},\mbox{Right}\}$ in red and $\{\mbox{Gray},\mbox{White}\}$ in blue. The coordinates are given by
$x=\min(a,c)/\max(a,c)$ and $y=b/\max(a,c)$, where $B=[a,b;b,c]$ is the observed block connectivity probability matrix. The thin black curve $y=\sqrt{x}$ represents the rank 1 submodel separating positive definite (lower right) from indefinite (upper left). The background color shading is Chernoff ratio $\rho$, and the thick black curves are $\rho=1$ separating the region where ASE is preferred (between the curves) from LSE preferred.
The point $(1,1)$ represents Erd\H{o}s-R\'enyi ($a=b=c$). The large stars are from the a priori composite connectome projections (Figure \ref{fig:LRandGW}).
We see that the red $\{\mbox{Left},\mbox{Right}\}$ projections are in the affinity region where $\rho < 1$ and LSE is preferred while the blue $\{\mbox{Gray},\mbox{White}\}$ projections are in the core-periphery region where $\rho > 1$ and ASE is preferred.
This analytical finding based on projections onto the SBM carries over to empirical spectral clustering results on the individual connectomes (Figure \ref{fig:BNU1DeltaARI}).
}
\label{fig:BNU1AffCP}
\end{figure}

\begin{figure}
\centering
\includegraphics[width=1.0\linewidth]{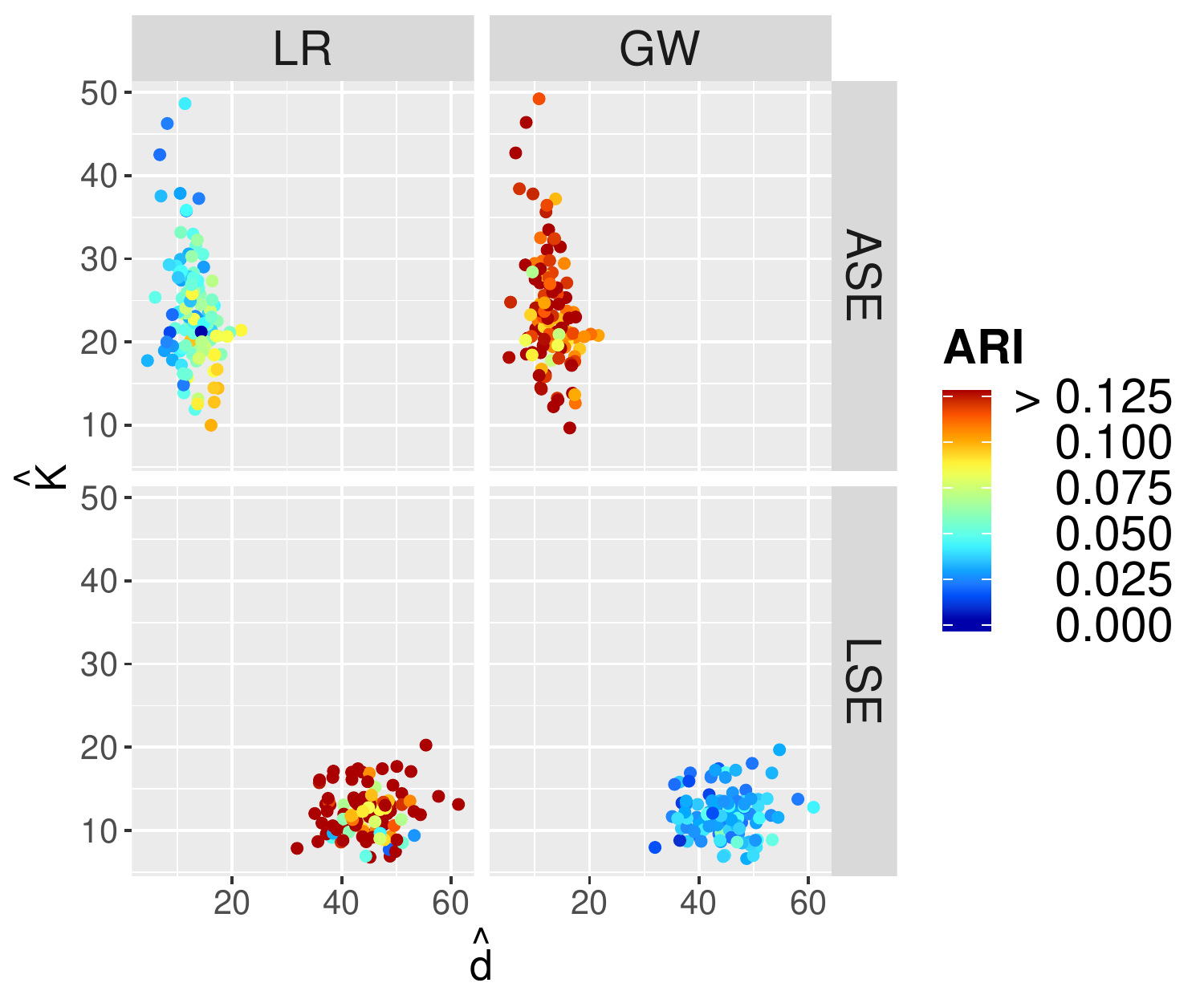}
\caption{
Results of the ($\widehat{d},\widehat{K}$) model selection for spectral graph clustering for each of our 114 connectomes.
For LSE we see
$\widehat{d} \in \{30,\dots,60\}$ and
$\widehat{K} \in \{2,\dots,20\}$;
for ASE we see
$\widehat{d} \in \{2,\dots,20\}$ and
$\widehat{K} \in \{10,\dots,50\}$.
The color-coding represents clustering performance in terms of ARI for each of LSE and ASE against each of the two truths \{Left,Right\} and \{Gray,White\},
and shows that LSE clustering identifies $\{\mbox{Left},\mbox{Right}\}$ better than $\{\mbox{Gray},\mbox{White}\}$ and ASE identifies $\{\mbox{Gray},\mbox{White}\}$ better than $\{\mbox{Left},\mbox{Right}\}$.
Our 'Two Truths' phenomenon is conclusively demonstrated: LSE finds $\{\mbox{Left},\mbox{Right}\}$ (affinity) while ASE finds $\{\mbox{Gray},\mbox{White}\}$ (core-periphery). 
}
\label{fig:xxx}
\end{figure}

\begin{figure}
\centering
\includegraphics[width=1.0\linewidth]{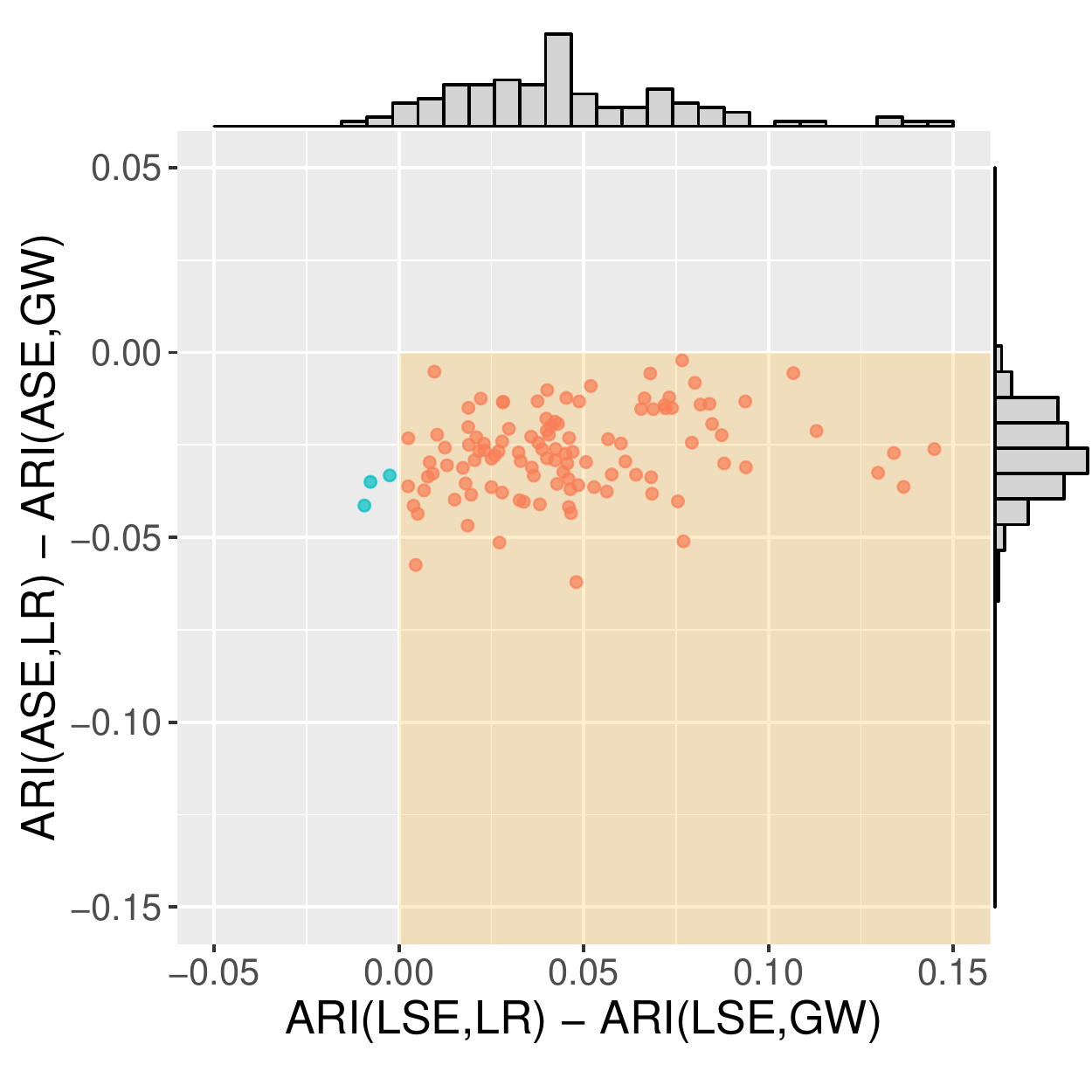}
\caption{Spectral graph clustering assessment via ARI. For each of our 114 connectomes, we plot the difference in ARI for the $\{\mbox{Left},\mbox{Right}\}$ truth against the difference in ARI for the $\{\mbox{Gray},\mbox{White}\}$ truth for the clusterings produced by each of LSE and ASE:
$x$ = ARI(LSE,LR) $-$ ARI(LSE,GW)
vs.\
$y$ = ARI(ASE,LR) $-$ ARI(ASE,GW).
A point in the $(+,-)$ quadrant indicates that for that connectome the LSE clustering identified $\{\mbox{Left},\mbox{Right}\}$ better than $\{\mbox{Gray},\mbox{White}\}$ and ASE identified $\{\mbox{Gray},\mbox{White}\}$ better than $\{\mbox{Left},\mbox{Right}\}$.
Marginal histograms are provided.
Our `Two Truths' phenomenon is conclusively demonstrated: LSE identifies $\{\mbox{Left},\mbox{Right}\}$ (affinity) while ASE identifies $\{\mbox{Gray},\mbox{White}\}$ (core-periphery).}
\label{fig:BNU1DeltaARI}
\end{figure}

Figures \ref{fig:BNU1AffCP}, \ref{fig:xxx} and \ref{fig:BNU1DeltaARI} present empirical results for the connectome data set, $m=114$ graphs each on $n \approx 40,000$ vertices.
We note that these connectomes are most assuredly {\emph not} 4-block `Two Truths' SBMs of the kind presented in Figures \ref{fig:LRGW} and \ref{fig:LRandGW}, but they do have `Two Truths' (\{Left,Right\} \& \{Gray,White\}) and, as we shall see, they do exhibit a real-data version of the synthetic results presented above, in the spirit of semiparametric SBM fitting.

First, 
in Figure \ref{fig:BNU1AffCP}, we consider a priori projections of the individual connectomes,
analogous to the Figure \ref{fig:LRandGW} projections of the composite connectome.
Letting $B=[a,b;b,c]$ be the observed block connectivity probability matrix  for the a priori 2-block SBM projection (\{Left,Right\} or \{Gray,White\}) of a given individual connectome,
the coordinates in Figure \ref{fig:BNU1AffCP} are given by
$x=\min(a,c)/\max(a,c)$ and $y=b/\max(a,c)$.
Each graph yields two points, one for each of 
\{Left,Right\} and \{Gray,White\}.
We see that the $\{\mbox{Left},\mbox{Right}\}$ projections are in the affinity region (large $x$ and small $y$ implies $a \approx c \gg b$, where Chernoff ratio $\rho < 1$ and LSE is preferred) while the $\{\mbox{Gray},\mbox{White}\}$ projections are in the core-periphery region (small $x$ and small $y$ implies $\max(a,c) \gg b \approx \min(a,c)$ where $\rho > 1$ and ASE is preferred).
This exploratory data analysis finding indicates complex `Two Truths' structure in our connectome data set.
(Of independent interest: we propose Figure \ref{fig:BNU1AffCP} as
the representative for a 
novel and illustrative 
`Two Truths' exploratory data analysis (EDA) plot
 for a data set of $m$ graphs with multiple categorical vertex labels.) 

In Figures \ref{fig:xxx} and \ref{fig:BNU1DeltaARI} we present the results of $m=114$ runs of the spectral clustering algorithm 
$\mbox{GMM} ~ \circ ~ \{\mbox{LSE},\mbox{ASE}\}$.
We consider each of LSE and ASE, 
choosing $\widehat{d}$ and $\widehat{K}$ as described above.
The resulting empirical clusterings are evaluated via ARI against each of the
\{Left,Right\} and \{Gray,White\} truths.
In Figure \ref{fig:xxx} we present the results of the ($\widehat{d},\widehat{K}$) model selection, and we observe that ASE is choosing $\widehat{d} \in \{2,\dots,20\}$ and LSE is choosing $\widehat{d} \in \{30,\dots,60\}$, while ASE is choosing $\widehat{K} \in \{10,\dots,50\}$ and LSE is choosing $\widehat{K} \in \{2,\dots,20\}$. 
In Figure \ref{fig:BNU1DeltaARI},
each graph is represented by a single point, plotting
$x$ = ARI(LSE,LR) $-$ ARI(LSE,GW)
vs.\
$y$ = ARI(ASE,LR) $-$ ARI(ASE,GW),
where ``LSE'' (resp.\ ``ASE'') represents the empirical clustering $\mathcal{C}_{LSE}$ (resp.\ $\mathcal{C}_{ASE}$) and ``LR'' (resp.\ ``GW'')
represents the true clustering
$\mathcal{C}_{\mbox{\small{\{Left,Right\}}}}$
(resp.\
$\mathcal{C}_{\mbox{\small{\{Gray,White\}}}}$).
We see that almost all of the points lie in the $(+,-)$ quadrant, indicating 
ARI(LSE,LR) > ARI(LSE,GW)
and
ARI(ASE,LR) < ARI(ASE,GW).
That is,
LSE finds the affinity \{Left,Right\} structure
and
ASE finds the core-periphery \{Gray,White\} structure.
The `Two Truths' structure in our connectome data set illustrated in Figure \ref{fig:BNU1AffCP}
leads to fundamentally different but equally meaningful LSE vs.\ ASE spectral clustering performance.
This is our `Two Truths' phenomenon in spectral graph clustering.

\newpage

\section*{Conclusion}


The results presented herein demonstrate that practical spectral graph clustering exhibits a `Two Truths' phenomenon with respect to Laplacian vs.\ Adjacency spectral embedding.
This phenomenon can be understood theoretically from the perspective of affinity vs.\ core-periphery Stochastic Block Models, and via consideration of the two a priori projections of a 4-block `Two-Truths' SBM onto the 2-block SBM.
For connectomics, this phenomenon manifests itself via
LSE better capturing the left hemisphere/right hemisphere affinity structure
and
ASE better capturing the gray matter/white matter core-periphery structure,
and suggests that a connectivity-based parcellation based on spectral clustering should consider both LSE and ASE,
as the two spectral embedding approaches facilitate the identification of different and complementary connectivity-based clustering truths.

\acknow{
This work is partially supported by DARPA (XDATA, GRAPHS, SIMPLEX, D3M), JHU HLTCOE, and the Acheson J.\ Duncan Fund for the Advancement of Research in Statistics.
The authors thank 
the Isaac Newton Institute for Mathematical Sciences,
Cambridge, UK, for support and hospitality during the programme Theoretical Foundations
for Statistical Network Analysis (EPSRC grant no.\ EP/K032208/1) where a portion of the
work on this paper was undertaken,
and
the University of Haifa, where these ideas were conceived in June 2014.

}

\showacknow{} 


\bibliography{pnas-sample}

\begin{thebibliography}{10}

\bibitem{Peele1602548}
Peel L, Larremore DB, Clauset A (2017) The ground truth about metadata and
  community detection in networks.
\newblock {\em Science Advances} 3(5).

\bibitem{vonLuxburg2007}
von Luxburg U (2007) A tutorial on spectral clustering.
\newblock {\em Statistics and Computing} 17(4):395--416.

\bibitem{rohe2011}
Rohe K, Chatterjee S, Yu B (2011) Spectral clustering and the high-dimensional
  stochastic blockmodel.
\newblock {\em Ann. Statist.} 39(4):1878--1915.

\bibitem{holland1983stochastic}
Holland PW, Laskey KB, Leinhardt S (1983) Stochastic blockmodels: First steps.
\newblock {\em Social networks} 5(2):109--137.

\bibitem{Olhede14722}
Olhede SC, Wolfe PJ (2014) Network histograms and universality of blockmodel
  approximation.
\newblock {\em Proceedings of the National Academy of Sciences}
  111(41):14722--14727.

\bibitem{athreya2013limit}
Athreya A, et~al. (2016) A limit theorem for scaled eigenvectors of random dot
  product graphs.
\newblock {\em Sankhya A} 78:1--18.

\bibitem{tang_lse}
Tang M, Priebe CE (2018) Limit theorems for eigenvectors of the normalized
  laplacian for random graphs.
\newblock {\em Annals of Statistics} 46:2360--2415.

\bibitem{PRD-GRDPG}
Rubin-Delanchy P, Priebe CE, Tang M, Cape J (2018) The generalised random dot
  product graph.
\newblock {\em https://arxiv.org/abs/1709.05506}.

\bibitem{chernoff_1952}
Chernoff H (1952) A measure of asymptotic efficiency for tests of a hypothesis
  based on the sum of observations.
\newblock {\em Annals of Mathematical Statistics} 23:493--507.

\bibitem{chernoff_1956}
Chernoff H (1956) Large sample theory: Parametric case.
\newblock {\em Annals of Mathematical Statistics} 27:1--22.

\bibitem{JCapeCC}
Cape J, Tang M, Priebe CE (2018) On spectral embedding performance and
  elucidating network structure in stochastic block model graphs.
\newblock {\em https://arxiv.org/abs/1808.04855}.

\bibitem{hubert85}
Hubert L, Arabie P (1985) Comparing partitions.
\newblock {\em Journal of Classification} 2(1):193--218.

\bibitem{dadiduar05}
Danon L, D{\'i}az-Guilera A, Duch J, Arena A (2005) Comparing community
  structure identification.
\newblock {\em Journal of Statistical Mechanics: Theory and Experiment}
  2005(09):P09008.

\bibitem{me07}
Meil{\u{a}} M (2007) Comparing clusterings--an information based distance.
\newblock {\em Journal of Multivariate Analysis} pp. 873--195.

\bibitem{ja1912}
Jaccard P (1912) The distribution of the flora in the alpine zone.
\newblock {\em The New Phytologist} 11(2):37--50.

\bibitem{Jackson}
Jackson JE (2004) {\em A User's Guide to Principal Components}.
\newblock (John Wiley \& Sons, Inc.).

\bibitem{chatterjee2015}
Chatterjee S (2015) Matrix estimation by universal singular value thresholding.
\newblock {\em The Annals of Statistics} 43(1):177--214.

\bibitem{Zhu:2006fv}
Zhu M, Ghodsi A (2006) Automatic dimensionality selection from the scree plot
  via the use of profile likelihood.
\newblock {\em Computational Statistics and Data Analysis} 51(2):918--930.

\bibitem{akaike1974new}
Akaike H (1974) A new look at the statistical model identification.
\newblock {\em IEEE Transactions on Automatic Control} 19(6):716--723.

\bibitem{BIC}
Schwarz G (1978) Estimating the dimension of a model.
\newblock {\em The Annals of Statistics} 6(2):461--464.

\bibitem{MDL}
Rissanen J (1978) Modeling by shortest data description.
\newblock {\em Automatica} 14(5):465 -- 471.

\bibitem{mclust2012}
Fraley C, Raftery AE (2002) Model-based clustering, discriminant analysis and
  density estimation.
\newblock {\em Journal of the American Statistical Association} 97:611--631.

\bibitem{Kiar188706}
Kiar G, et~al. (2018) A high-throughput pipeline identifies robust connectomes
  but troublesome variability.
\newblock {\em https://www.biorxiv.org/node/94401}.

\end{thebibliography}


\end{document}